# Detailed Technical Report

as part of participation in

# Higgs Boson Machine Learning Challenge

Organized by
Kaggle, CERN, ATLAS et al

Prepared for:
Prof. Sultan Sial

Prepared by:
S. Raza Ahmad
Research Assistant
Department of Mathematics
Lahore University of Management Sciences
(LUMS), Lahore

# Table of Contents

# Abstract

This report entails the detailed description of the approach and methodologies taken as part of competing in the Higgs Boson Machine Learning Competition hosted by Kaggle Inc. and organized by CERN et al. It briefly describes the theoretical background of the problem and the motivation for taking part in the competition. Furthermore, the various machine learning models and algorithms analyzed and implemented during the 4 month period of participation are discussed and compared. Special attention is paid to the Deep Learning techniques and architectures implemented from scratch using Python and NumPy for this competition.

# Introduction

**Physics Background:**

The discovery of Higgs particle was announced on 4$^{th}$ July 2012. In 2013, Nobel Prize was conferred upon two scientists, Francois Englert and Peter Higgs for their contribution towards its discovery. A characteristic property of Higgs Boson is its decay into other particles through different processes.

At the ATLAS detector at CERN, very high energy protons are accelerated in a circular trajectory in both directions thus colliding with themselves and resulting in hundreds of particles per second. These events are categorized as either background or signal events. The background events consist of decay of particles that have already been discovered in previous experiments. The signal events are the decay of exotic particles: a region in feature space which is not explained by the background processes. The significance of these signal events is analyzed using different statistical tests. If the probability that the event has not been produced by a background process is well below a threshold, a new particle is considered to have been discovered. The ATLAS experiment observed a signal of the Higgs Boson decaying into two tau particles, although it was buried in significant amount of noise.

**Machine Learning Background:**

High energy physicists use different machine learning techniques to optimize and investigate the selection region that produces these signal events. Classifiers are trained on simulated signal and background events that are assigned a weight to compensate for the discrepancy between the prior probability of the event and the probability in the simulator.

The Higgs Boson Machine Learning Challenge commenced on 14$^{th}$ May 2014 and culminated on 15$^{th}$ September 2014. The goal of the challenge was to improve the techniques that produce the signal selection region. A formal objective function was introduced called Approximation of the Median Significance (AMS). It was a function of the weights of selected events which took into account the unnormalized true and false positives rates. The problem is formally defined and elaborated in the technical documentation provided by the organizers and could be accessed here.
*http://higgsml.lal.in2p3.fr/documentation/*

**Data:**

The data consists of simulated signal and background events in a 30 dimensional feature space. Each event data point is assigned an ID and a weight as explained before. The 30 features consisted of real values and included different kinematic properties of that event and the particles involved including estimated particle mass, invariant mass of hadronic tau and lepton, vector sum of the transverse momentum of hadronic tau, centrality of azimuthal angle, pseudo-rapidity of the leptons, the number of jets and their properties, etc. The training data consisted of 250,000 events and the test data consisted of 550,000 events. Test data was not accompanied by weights. Each event of training data was marked by one of two labels; 's' for signal and 'b' for background. The task was to submit a file consisting of 's' and 'b' predictions for each point in the test set.

Values of some feature were invalid for some of the events; those features or properties were missing in those events. Such values were marked -999.0 which was quite out of the range of other normal feature values.

# Initial Implementations

The baseline submission was made by the starting toolkit made available by the organizers on Kaggle. It was implemented using Binned Naive Bayes. The other initial solutions aimed at classification of data were implemented using open source scikit-learn library in Python. The total number of models used were approximately 12 in number. Following is a list of these models.

1. KMeans Clustering
2. Affinity Propagation
3. Spectral Clustering
4. K-Nearest Neighbors
5. AdaBoost Classifier
6. Bagging Classifier
7. Random Forest Classifier
8. Gradient Boosting Classifier
9. Gaussian Mixture Models
10. Naive Bayes Classifier
11. Support Vector Machines
12. Decision Tree Classifier

The maximum AMS score on public leader board achieved by these classifiers -and overall- was 3.38 by Gradient Boosting Classifier with threshold cutoff value of 85.5. The classification accuracy it achieved was 84%. The details of hyper parameter settings for this best submission are as follows.

- n_estimators = 100
- max_depth = 5
- min_samples_leaf = 200
- max_features = 10
- learning_rate = 0.5

Note: Detailed description of these features could be found in the scikit-learn API on their website www.scikit-learn.org.

The other classifiers did not prove to be any greater in accuracy. Support vector machines with linear kernels managed an accuracy of 76% and those with radial basis kernels proved intractable in performing predictions. K nearest neighbors algorithm using K = {3, 5, 7, 9} was executed but it did not improve upon the accuracy of support vector machines. The implementation of 9NN ran for about 2 days and even scored less than the prediction accuracy of SVM's.

The interesting hack tried was the use of a mixture of supervised and unsupervised algorithms. As mentioned earlier, some of the feature values were invalid for some events and were hence marked -999.0. It was observed that the there were a total of 6 different possible combinations of events that had one or more of the features marked invalid. The invalid features could altogether be eliminated when the events with the same configuration of invalid features were considered as independent training sets. This gave rise to 6 mini-training sets which could be trained independently. At test time, the test events were again divided on the basis of different configurations of these invalid features and sent to the appropriate classifier for prediction. This technique did not prove very helpful in classification, possible because of highly skewed mini-training sets.

Another small hack was the use of unsupervised learning using KMeans Clustering, Gaussian Mixture Models and Affinity Propagation. It was expected that since there were 4 distinct distributions of data within the dataset; 1 signal data and 3 different kinds of background data, the algorithms would try to cluster together points from the same class. The technique did not work out, perchance due to skewed clusters and the absence of high affinity among the data points of the same original distribution.

# Deep Learning Implementation

**Introduction:**

Deep learning is a new area in Machine Learning that attempts to model high level abstractions present in the raw data to understand the high varying functions underlying the data and to perform well generalized predictions for unseen data. This is accomplished through certain non-linear transformations of data through varying deep architectures such as Neural Networks. Deep learning aims at fulfilling the objective of true Artificial Intelligence and has recently been of great interest to researchers in machine learning. Tech giants like Google, Microsoft, Facebook and Baidu are investing hundreds of millions of dollars in bleeding-edge deep learning research and developing its applications.

The main focus of this research work remained on studying and implementing different deep learning techniques for searching high energy exotic particles in this competition. The deep networks employed were different kinds of deep neural networks. The motivation partly came from the works of P. Baldi et al of UC Irvine who had recently worked on Higgs and Susy datasets with significantly improved results over the benchmark machine learning techniques as the likes of Boosted Decision Trees. Other motivations came from the works of Geoffery Hinton of University of Toronto, Yoshua Bengio of University of Montreal, Yann LeCun of New York University and Andrew Ng of Stanford University.

**Architecture:**

The deep architecture constructed for the competition was a deep feed forward neural network with a total of 5 layers; 4 hidden and 1 output. Output layer consisted of one whereas each hidden layer consisted of 300 logistic units. All of the units used the sigmoid activation function. Linear and tanh activation functions were also tried but did yield good enough results.

Apart from the prime architecture, many other different deep and shallow neural networks were also designed and compared with. These consisted of networks having 2, 3, 4 and 6 layers with varying number of units in each layer.

## Implementation:

Implementation was done in Python using NumPy and SciPy open source libraries and run on a distributed cluster of 12 nodes running Red Hat Enterprise Linux having Xeon processors and a memory of 64 GB(Rustam3). Parameter optimization was performed using Stochastic Gradient Descent with mini-batches of size 50. Initial learning rate and momentum were set to 0.05 and 0.9 respectively. Training ended when number of epochs reached a maximum of 500 and the minimum error on a 20% held-out validation set did not decrease for the last 30 epochs by a factor of 0.001.

The learning rate decreased through an annealing schedule by a factor of 0.0005 every epoch. Momentum increased linearly over the first 100 epochs from 0.9 to 0.99 and then remained constant. RMSProp technique was used with a beta value of 0.9. Weights were initialized from a Gaussian distribution having zero mean and a variance of 0.1 in the first layer, 0.05 in rest of the hidden layers and 0.01 in the output layer. All the hidden layers were pre-trained using Greedy layer wise pre-training algorithm using stacked auto-encoders each having 1 hidden layer. Hyper parameters for the network were optimized using different subsets of training data of sizes 1000, 10000 and 50000.

## Results:

The limited time allocated for implementation resulted in results not quite as good comparable to the gradient boosting classifier. The scripts ran for about 7-10 days and the best accuracy achieved was 83% and the AMS 2.1. This was done by averaging together different kinds of deep models.

# Conclusion

In the aftermath of participation in the competition, a hands-on machine learning experience in a real life competitive setting has been achieved. This would surely prove helpful in the works to follow. There is a need to learn more about different kinds of learning models and their underly theoretical background.

**Important Considerations:**

The following things are just a few that should be kept in mind for future research endeavors.

1. One month time is usually not sufficient for the implementation of a project. Implementation should start in the very beginning and literature review should be done side by side.
2. A methodical approach should be taken about which techniques to try during the project keeping in view its scope and bounds.
3. A proper plan should be carved out on how to implement the code and either any libraries should be employed or a custom code be written from scratch.
4. A complete record should be maintained for each implementation in an automated manner. Version control system should be adopted.
5. Routines for automated testing should be created on the way and unit tests be performed after every update.
6. Object oriented programming should be applied to possibly each and every component of the code. Tweaking hyper parameters of the model should just incur the cost of changing a few values.
7. One model or model averaging is not always the best possible thing. Ensemble methods should be used and boosting should be employed.
8. Feature engineering and extraction should be considered more seriously even with deep nets.

# Appendix

I. Details and variance of features:

| Attribute | Minimum | Maximum | Mean | Standard Deviation\ | Unique Values |
|---|---|---|---|---|---|
| DER_mass_MMC | -999 | 1192.026 | -49.02307944 | 406.344834011 | 108338 |
| DER_mass_transverse_met_lep | 0 | 690.075 | 49.239819276 | 35.344814922 | 101637 |
| DER_mass_vis | 6.329 | 1349.351 | 81.181981612 | 40.828608875 | 100558 |
| DER_pt_h | 0 | 2834.999 | 57.895961656 | 63.6555543068 | 115563 |
| DER_deltaeta_jet_jet | -999 | 8.503 | -708.4206754 | 454.479656149 | 7087 |
| DER_mass_jet_jet | -999 | 4974.979 | -601.237050732 | 657.970986168 | 68366 |
| DER_prodeta_jet_jet | -999 | 16.69 | -709.3566029 | 453.018970512 | 16593 |
| DER_deltar_tau_lep | 0.208 | 5.684 | 2.373099844 | 0.7829095528 | 4692 |
| DER_pt_tot | 0 | 2834.999 | 18.917332444 | 22.2734492049 | 59042 |
| DER_sum_pt | 46.104 | 1852.462 | 158.432217048 | 115.705883721 | 156098 |
| DER_pt_ratio_lep_tau | 0.047 | 19.773 | 1.437609432 | 0.8447412552 | 5931 |
| DER_met_phi_centrality | -1.414 | 1.414 | -0.128304708 | 1.1935824486 | 2829 |
| DER_lep_eta_centrality | -999 | 1 | -708.985189132 | 453.595814008 | 1002 |
| PRI_tau_pt | 20 | 764.408 | 38.707419128 | 22.4120358425 | 59639 |
| PRI_tau_eta | -2.499 | 2.497 | -0.010973048 | 1.2140762179 | 4971 |
| PRI_tau_phi | -3.142 | 3.142 | -0.008171072 | 1.8167594109 | 6285 |
| PRI_lep_pt | 26 | 560.271 | 46.660207248 | 22.0648782751 | 61929 |
| PRI_lep_eta | -2.505 | 2.503 | -0.019507468 | 1.2649796185 | 4987 |
| PRI_lep_phi | -3.142 | 3.142 | 0.043542964 | 1.8166076296 | 6285 |
| PRI_met | 0.109 | 2842.617 | 41.717234524 | 32.8946274025 | 87836 |
| PRI_met_phi | -3.142 | 3.142 | -0.010119192 | 1.8122190775 | 6285 |
| PRI_met_sumet | 13.678 | 2003.976 | 209.797177632 | 126.499252717 | 179740 |
| PRI_jet_num | 0 | 3 | 0.979176 | 0.9774243505 | 4 |
| PRI_jet_leading_pt | -999 | 1120.573 | -348.329567188 | 532.961723432 | 86590 |
| PRI_jet_leading_eta | -999 | 4.499 | -399.254313892 | 489.337307341 | 8558 |
| PRI_jet_leading_phi | -999 | 3.141 | -399.259788008 | 489.332904652 | 6285 |
| PRI_jet_subleading_pt | -999 | 721.456 | -692.381203548 | 479.874536093 | 42464 |
| PRI_jet_subleading_eta | -999 | 4.5 | -709.121609164 | 453.383717278 | 8628 |
| PRI_jet_subleading_phi | -999 | 3.142 | -709.118631136 | 453.388110495 | 6286 |
| PRI_jet_all_pt | 0 | 1633.433 | 73.064591384 | 98.0154659767 | 103559 |

II. Highest varying dimensions in data:
   (Using Principal Components Analysis)

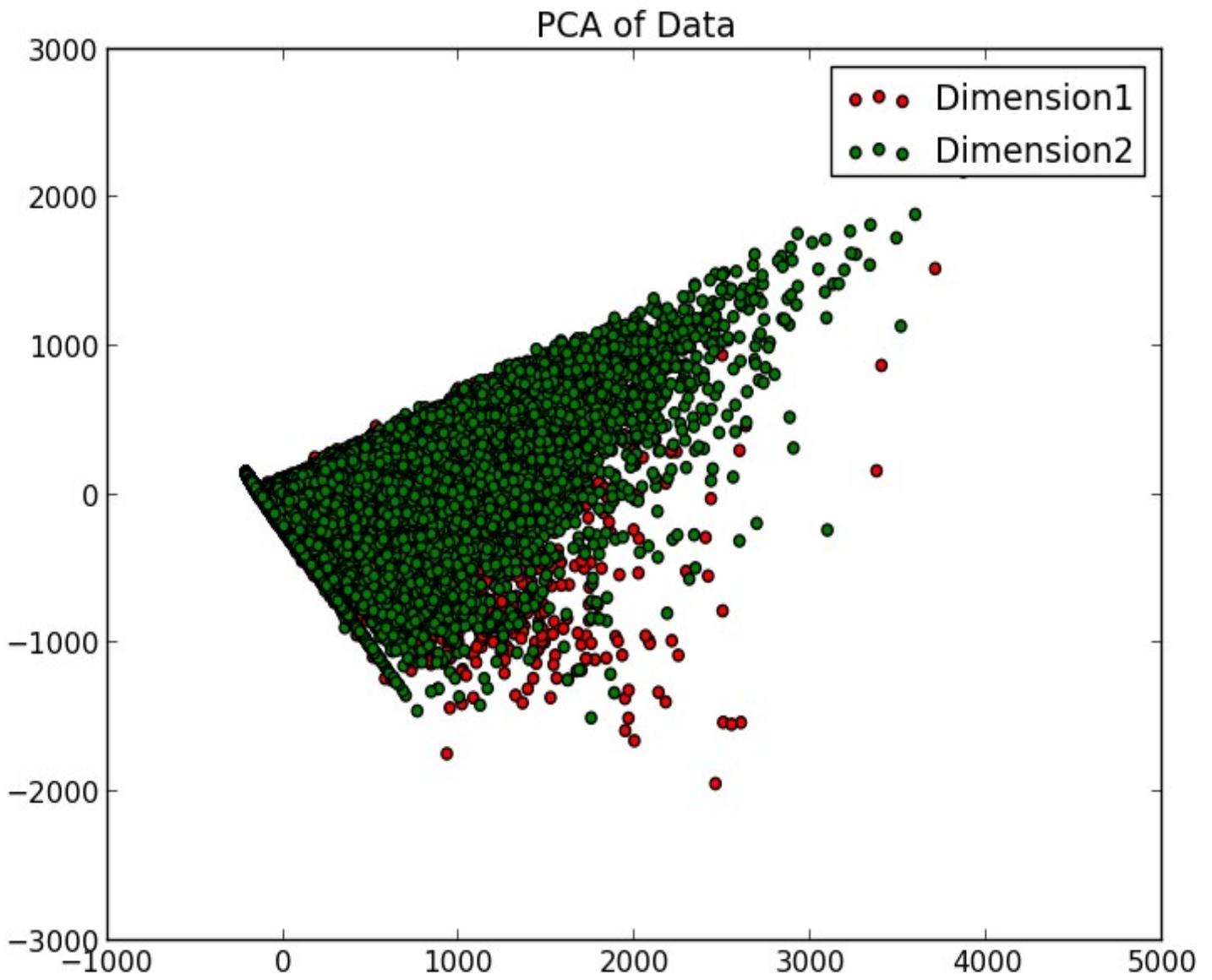

# Pointers

1. Challenge Website
   http://www.kaggle.com/c/higgs-boson/
2. Winning solution description
   http://www.kaggle.com/c/higgs-boson/forums/t/10425/code-release
3. Second place solution description
   https://github.com/TimSalimans/HiggsML
4. third place solution description
   http://www.kaggle.com/c/higgs-boson/forums/t/10481/third-place-model-documentation/55390#post55390